\pdfoutput=1
\documentclass[11pt]{article}

\usepackage[]{acl}

\usepackage{times}
\usepackage{latexsym}
\usepackage{float}
\usepackage{amsthm}
\usepackage{tcolorbox}

\usepackage[T1]{fontenc}
\usepackage[utf8]{inputenc}

\usepackage{microtype}
\usepackage{amsmath, amssymb}
\usepackage{booktabs}

\usepackage{graphicx}
\usepackage{multirow}

\usepackage{bm}
\usepackage{algorithm}
\usepackage{pifont}
\usepackage{subfigure}
\usepackage{xspace}

\title{Large Language Models are not Fair Evaluators}

\begin{document}

\author{
Peiyi Wang$^1$ \quad Lei Li$^1$ \quad Liang Chen$^1$ \quad Zefan Cai$^1$ \quad Dawei Zhu$^1$  \\
  \textbf{Binghuai Lin$^3$  \quad  
Yunbo Cao$^3$ \quad  Qi Liu$^2$ \quad Tianyu Liu$^3$ \quad Zhifang Sui$^1$}
\\ 
% $^1$ Key Laboratory of Computational Linguistics, Peking University, MOE, China \\
%
$^1$ National Key Laboratory for Multimedia Information Processing, Peking University\\
$^2$ The University of Hong Kong $^3$ Tencent Cloud AI \\
 \texttt{\{wangpeiyi9979, nlp.lilei, zefncai\}@gmail.com} \\
 \texttt{leo.liang.chen@outlook.com};\  \texttt{\{dwzhu, szf\}@pku.edu.cn} \\ \texttt{liuqi@cs.hku.hk}; \ \texttt{\{binghuailin, yunbocao, rogertyliu\}@tencent.com} 
}

\maketitle
\begin{abstract}
In this paper, we uncover a systematic bias in the evaluation paradigm of adopting large language models~(LLMs), e.g., GPT-4, as a referee to score and compare the quality of responses generated by candidate models.
We find that the quality ranking of candidate responses can be easily hacked by simply altering their order of appearance in the context. 
This manipulation allows us to skew the evaluation result, making one model appear considerably superior to the other, e.g., Vicuna-13B could beat ChatGPT on 66 over 80 tested queries with ChatGPT as an evaluator. 
To address this issue, we propose a calibration framework with three simple yet effective strategies:
1) Multiple Evidence Calibration, which requires the evaluator model to generate multiple evaluation evidence before assigning ratings; 
2) Balanced Position Calibration, which aggregates results across various orders to determine the final score;
3) Human-in-the-Loop Calibration, which introduces a balanced position diversity entropy to measure the difficulty of each example and seeks human assistance when needed.
We also manually annotate the ``win/tie/lose'' outcomes of responses from ChatGPT and Vicuna-13B in the Vicuna Benchmark's question prompt, and
extensive experiments demonstrate that our approach successfully mitigates evaluation bias, resulting in closer alignment with human judgments. 
We release our code and human annotation at \url{https://github.com/i-Eval/FairEval} to facilitate future research.
\end{abstract}

\section{Introduction}
The rapid advancement of Large Language Models (LLMs)~\citep{Brown:2020gpt3,Chowdhery2022PaLMSL} has underscored the importance of evaluating their alignment with human intent in generated responses, making it an active field of research. 
Traditional n-gram metrics like BLEU \cite{papineni-etal-2002-bleu} and ROUGE \cite{lin-2004-rouge}, as well as more sophisticated model-based evaluations such as BERTScore \cite{bertscore} and BARTScore \cite{bart-score}, are insufficient for thoroughly assessing this alignment \cite{he-etal-2023-blind}.
While human evaluation provides the most accurate measure of model performance and valuable insights, it can often be costly and time-consuming. As a result, there is a growing demand for automated assessment methods that can consistently align with human judgments while being more efficient and cost-effective.

\begin{figure}[t!]
    \centering
    \includegraphics[width=1\linewidth]{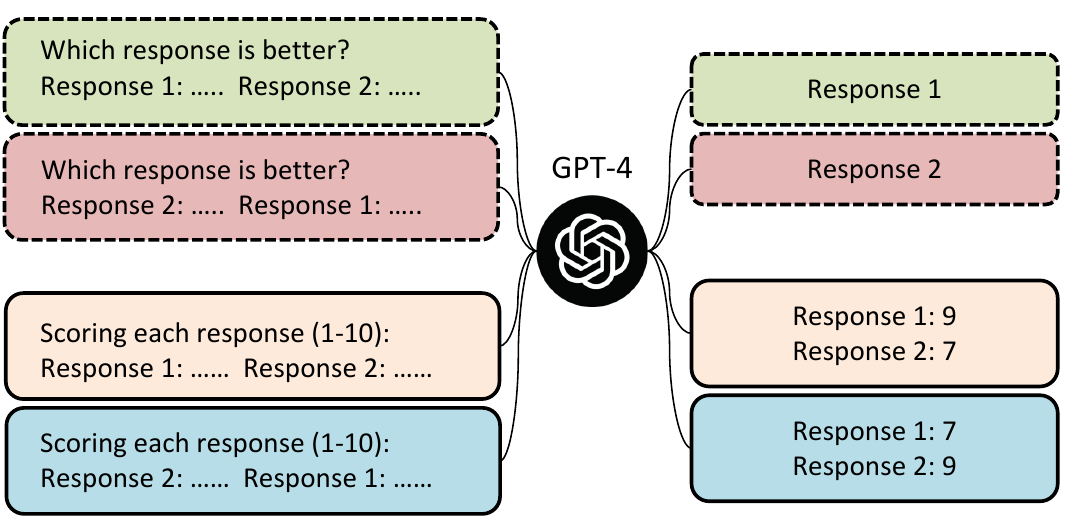}
    \caption{Simply changing the order of  candidate responses leads to overturned comparison results, even though we add the command ``ensuring that the order in which the responses were presented does not affect your judgment'' into the prompt.}
    \label{fig:intro_data}
\end{figure}

ChatGPT~\citep{chatgpt} and GPT-4 \cite{gpt4} have recently demonstrated remarkable performance across various tasks, leading to their widespread use as both the annotators \cite{it-with-gpt4, Xu2023BaizeAO} and evaluators \cite{vicuna2023, it-with-gpt4, Sun2023PrincipleDrivenSO, Zhou2023LIMALI, Gao2023LLaMAAdapterVP,PandaLM, dubois2023alpacafarm, wang2023far}. For example, The evaluation pipeline of Vicuna \cite{vicuna2023} has gained significant interest and wide usage due to its simplicity and interpretability. 
It prompts GPT-4 to score and compare candidate responses and provide explanations, making it a valuable tool for evaluation.
However, it is unclear how reliable LLMs are as evaluators, as they are known to be sensitive to textual instructions and inputs \cite{icl-survey,unfaithful-cot,bowman2023eight}. 
This raises questions about the resilience of this paradigm against perturbations, such as the ordering of candidates during scoring, potentially becoming the Achilles' Heel that can be easily hacked for unreliable evaluations.

In this paper, we take a sober look at the LLMs-as-evaluator paradigm and uncover a significant positional bias. Specifically, we demonstrate that GPT-4 exhibits a preference for the first displayed candidate response by consistently assigning it higher scores, even when the order of candidates is subtly altered. As illustrated in Figure \ref{fig:intro_data}, merely swapping the presentation order can reverse evaluation outcomes. This bias is also present in ChatGPT, which typically favors the second response. These findings highlight previously overlooked limitations in the current evaluation paradigm.

To address this issue, we propose three simple yet effective strategies to calibrate positional bias:
\textbf{1) Multiple Evidence Calibration (MEC)}: We prompt the model to generate evaluation evidence before assigning scores, leveraging the inherent properties of causal language models for calibration. We also employ ensemble techniques to incorporate multiple evidence calibration results to further stabilize the evaluation.
\textbf{2) Balanced Position Calibration (BPC)}: To further reduce positional bias, we evaluate each candidate in both positions across two runs and compute the final score as the average of the two runs.
\textbf{3) Human In The Loop Calibration (HITLC)}:
We also explore human-in-the-loop evaluation and consider a diversity-based method to get a cue to indicate biased candidates based on the evaluation results of MEC and BPC.

To assess the efficacy of our methods, we manually annotate the ``win/tie/lose'' outcomes of responses from ChatGPT and Vicuna-13B in the Vicuna benchmark \cite{vicuna2023}, encompassing $80$ questions spanning $9$ distinct question categories.
Our MEC and BPC enhance the evaluation alignment of GPT-4 and ChatGPT by $9.8$\% and $14.3$\% accuracy, respectively. 
Moreover, based on MEC and BPC, our HITLC can further effectively integrate human assistance into the evaluation process. 
Specifically, with only a $20$\% human annotation cost, GPT-4 and ChatGPT can achieve comparable or even better annotation alignment with the average human performance, reducing the annotation cost by up to $39$\%.

In summary, our key contributions are:
\textbf{1)} We reveal that LLMs exhibit severe positional bias, compromising their fairness as evaluators;
\textbf{2)} We develop a calibration framework with three simple yet effective strategies to calibrate the positional bias of LLMs;
\textbf{3)} We manually annotate the ``win/tie/lose'' outcomes of responses from ChatGPT and Vicuna-13B in the Vicuna benchmark and demonstrate the effectiveness of our proposed approach through experimental results, which show closer alignment with human judgments.

\begin{table}[t]
\small
\begin{tcolorbox}

[Question] 

\textcolor[rgb]{0,0,0.9}{\{Q\}}

[The Start of Assistant 1's response] 

\textcolor[rgb]{0,0,0.9}{\{R1\}}

[The End of Assistant 1's response]

[The Start of Assistant 2's response] 

\textcolor[rgb]{0,0,0.9}{\{R2\}}

[The End of Assistant 2's response] 

[System] 

We would like to request your feedback on the performance of two AI assistants in response to the user question displayed above. 

Please rate the helpfulness, relevance, accuracy, level of details of their responses. Each assistant receives an overall score on a scale of 1 to 10, where a higher score indicates better overall performance. 

Please first output a single line containing only two values indicating the scores for Assistant 1 and 2, respectively.

The two scores are separated by a space. In the subsequent line, please provide a comprehensive explanation of your evaluation, avoiding any potential bias and \textcolor[rgb]{0.8,0,0}{ensuring that the order in which the responses were presented does not affect your judgment.}

\end{tcolorbox}
\caption{The evaluation template with three slots (\textcolor[rgb]{0,0,0.9}{\{Q\}, \{R1\} and \{R2\}}) from \citet{vicuna2023}. Even though the template emphasizes not letting the order affect the results (\textcolor[rgb]{0.8,0,0}{red text}), large language models still have a large positional bias.}
\label{tab:template}
\end{table}

\begin{table*}[th]
\centering
\small
\begin{tabular}{llccc}
\toprule
\multirow{2}{*}{\textbf{\textsc{Evaluators}}}  & \multirow{2}{*}{\textbf{\textsc{Vicuna-13B v.s. Other Models}}}  &  \multicolumn{2}{c}{\textbf{\textsc{Vicuna-13B Win Rate}}}                   &  \multirow{2}{*}{\textbf{\textsc{Conflict Rate}}} \\ 
\cmidrule(r){3-4} 
                                       & &  \multicolumn{1}{c}{\textsc{as Assistant1}} & \multicolumn{1}{c}{\textsc{as Assistant2}} &                                      \\ \midrule
GPT-4 & Vicuna-13B v.s. ChatGPT          &  51.3\%    &  23.8\%    &  37 / 80 (46.3\%)                      \\ 
GPT-4 & Vicuna-13B v.s. Alpaca-13B             &  92.5\%     &  92.5\%     &  4 / 80 (5.0\%)                        \\
\midrule
ChatGPT & Vicuna-13B v.s. ChatGPT          &  2.5\%    &  82.5\%  &  66 / 80 (82.5\%)                      \\ 
ChatGPT & Vicuna-13B v.s. Alpaca-13B             &  37.5\%     &  90\%    &  42 / 80 (52.5\%)                        \\

\bottomrule
\end{tabular}
\caption{The Win Rate of Vicuna-13B significantly fluctuates when positioned as Assistant 1 and Assistant 2, with GPT-4 and ChatGPT serving as evaluators. \textsc{Conflict Rate} refers to the proportion of conflicting results given by the same evaluator when simply changing the position of two models.}
\label{tab:pilot}
\end{table*}

\section{Positional Bias of the LLM Evaluator}

\subsection{LLMs as Evaluators}

Recently, researchers have been utilizing LLMs such as GPT-4 as evaluators to compare the performance of two AI assistants. 
As shown in Table \ref{tab:template},  an evaluation template with three placeholders 
$T (Q, R1, R2)$, is used to query the LLM for evaluation.
For each testing question $q$, given two responses $r1$ and $r2$ from Assistant 1 and Assistant 2, respectively, the researchers populate these responses into the corresponding slots of the evaluation template to form a prompt:
$T (Q=q, R1=r1, R2=r2).$
The prompt is then used to query the LLM in order to obtain the comparison result.
In this paper, we found that LLM suffers from severe positional bias, i.e., by swapping the slots of the two responses and querying LLM twice, the evaluator will most likely produce conflicting evaluation results, and the evaluator prefers the response at a certain position.

\subsection{Revealing the Positional Bias}

In this section, we adopt GPT-4 and ChatGPT as evaluators to analyze the characteristics of positional bias in LLM evaluators. 
We find that:

\paragraph{LLMs are sensitive to the position of responses.}
As shown in Table \ref{tab:pilot},
in the evaluation of ``{Vicuna-13B v.s. ChatGPT}'' and ``{Vicuna-13B v.s. Alpaca-13B}'', when the order was changed, LLMs provide different evaluation results, e.g., the win rate of Vicuna-13B extremely differs when Vicuna-13B is evaluated as Assistant 1 and Assistant 2.

To empirically evaluate the sensitivity, we introduced a metric  $\mathbf{Conflict \ Rate}$ to measure the sensitivity of the model to response positions quantitatively.
Formally, given $N$ examples  $\{(q_i, r1_i, r2_i)\}_{i=1}^N$,
for each example $(q_i, r1_i, r2_i)$, we query the LLM with two prompts $T(q_i, r1_i, r2_i)$ and $T(q_i, r2_i, r1_i)$, and obtain corresponding two evaluation results $\mathbf{ER}_i^{r12}$ and $\mathbf{ER}_i^{r21}$.
Then we calculate the Conflict Rate of the LLM evaluator as follows:
\begin{equation}
\mathbf{Conflict \ Rate} = \frac{\sum_{i=1}^{N} \mathbb{I}   (\mathbf{ER}_i^{r12} \neq \mathbf{ER}_i^{r21})}{N},
\end{equation}
where $ \mathbb{I(.)} $ is the indicator function. 
We found that GPT-4 exhibited conflict rates of 46.3\% and 5.0\%, respectively. 
In contrast, ChatGPT displayed considerably higher conflict rates, with figures of 82.5\% and 52.5\%, respectively.
These findings indicate that LLMs can be self-conflicting due to the sensitivity of the response order in the template, with stronger models being less influenced by the placement of responses.

\paragraph{LLMs suffer from Positional Bias, i.e., they prefer the response in the specific position.}
Based on the same evaluation template $T$ in Table \ref{tab:template}, GPT-4 tends to favor the response in the first position, while ChatGPT shows a preference for the response in the second position.
For example, as illustrated in Table \ref{tab:pilot}, in the comparison ``{Vicuna-13B v.s. ChatGPT}'', GPT-4 yields Win Rates of $51.3$\% and $23.8$\% for Vicuna-13B when it is positioned as Assistant 1 and Assistant 2, respectively. 
Conversely, ChatGPT indicates Win Rates of only $2.5$\% and up to $82.5$\% for Vicuna-13B when it is positioned as Assistant 1 and Assistant 2, respectively.

\begin{figure}[t]
    \centering
    \includegraphics[width=0.95\linewidth]{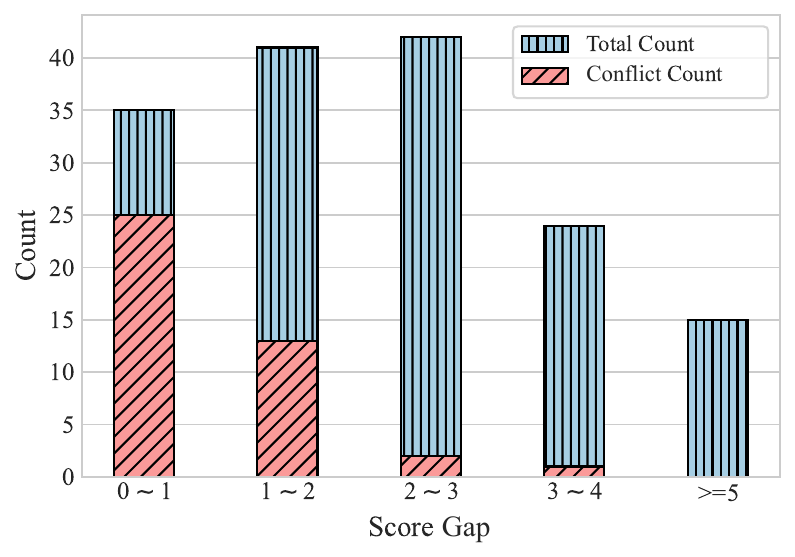}
    \caption{The conflict rate is negatively correlated with the score gap between the two responses. When swapping the order of two responses, the smaller the score gap between them, the more likely GPT-4 is to produce conflicting results.}
    \label{fig:bin}
\end{figure}

\begin{table}[t]
\small
\centering
\scalebox{1}{
\begin{tcolorbox}

[Question] 

\textcolor[rgb]{0,0,0.9}{\{Q\}}

[The Start of Assistant 1's response] 

\textcolor[rgb]{0,0,0.9}{\{R1\}}

[The End of Assistant 1's response]

[The Start of Assistant 2's response] 

\textcolor[rgb]{0,0,0.9}{\{R2\}}

[The End of Assistant 2's response]

[System]

We would like to request your feedback on the performance of two AI assistants in response to the user question displayed above.

Please rate the helpfulness, relevance, accuracy, and level of detail of their responses. 
Each assistant receives an overall score on a scale of 1 to 10, where a higher score indicates better overall performance.

\textcolor[rgb]{0.8,0,0}{Please first provide a comprehensive explanation of your evaluation}, avoiding any potential bias and ensuring that the order in which the responses were presented does not affect your judgment. Then, output two lines indicating the scores for Assistant 1 and 2, respectively.

Output with the following format:

Evaluation evidence: <evaluation explanation here>

The score of Assistant 1: <score>

The score of Assistant 2: <score>

\end{tcolorbox}}
\caption{The evidence calibration evaluation template that prompts LLMs to generate the evaluation evidence first (\textcolor[rgb]{0.8,0,0}{red text}), and then evaluate two responses. }
\label{tab:template_cot}
\end{table}

\begin{figure*}[t]
    \centering
    \includegraphics[width=1\linewidth]{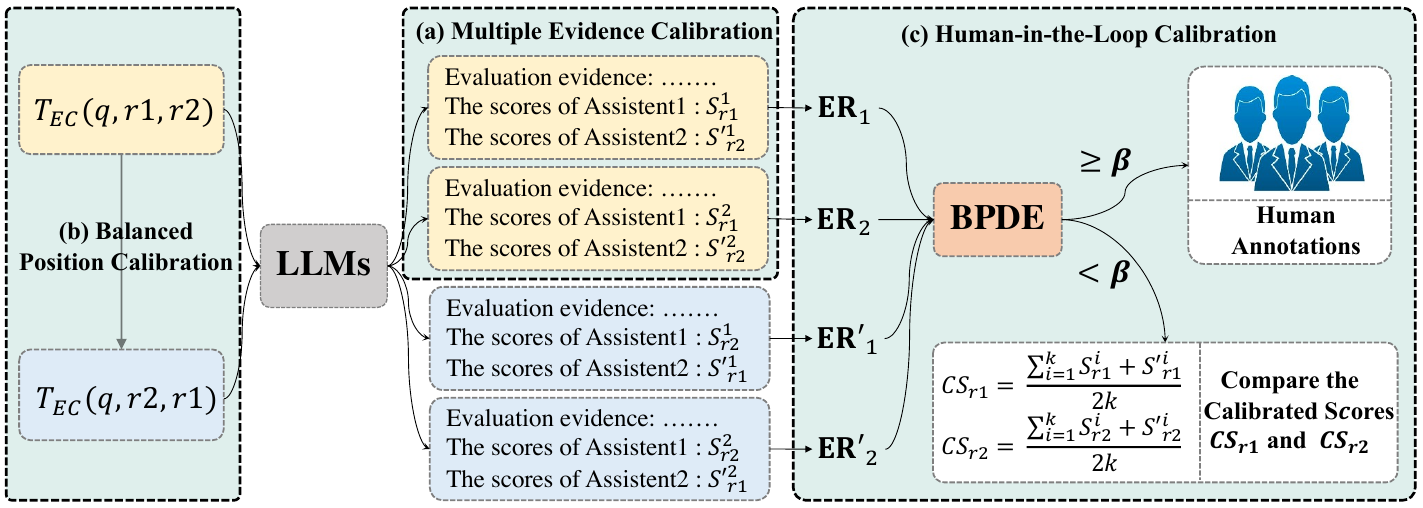}
    \caption{Demonstration of our calibration framework with three calibration methods. $S_r$ and $S_{r}^{'}$ denotes scores of the response $r$ in the first and second positions, respectively. BPDE is short for Balanced Position Diversity Entropy score, which is calculated based on the evaluation results (ER) of MEC and BPC.}
    \label{fig:method}
\end{figure*}

\paragraph{The degree of positional bias varies based on the difference in response quality.}
\label{sec:dif-bias}
We notice that the conflict rate of ``{Vicuna-13B v.s. Alpaca-13B}'' is much lower than that of ``{Vicuna-13B v.s. ChatGPT}'',  suggesting that positional bias may not have the same impact on the assessment of different responses.
One potential reason is that there is a significant difference in the quality of responses between Alpaca models and Vicuna models, and positional bias is not strong enough to change the judgment in such a situation.
To further investigate this issue, we grouped all the examples based on the score difference between the two responses.
As shown in Figure \ref{fig:bin}, we found that when the score difference between the two responses is small (e.g., score gap $\le$ 1), the evaluation results of GPT-4 are significantly affected by the position of the responses. On the other hand, when the score difference between the two responses is large (e.g., score gap $\ge$ 3), GPT-4's evaluation results are relatively stable.

\section{Calibrating the Positional Bias}
We have identified that positional bias can significantly impact the evaluation results of LLMs, making them unfair evaluators.
In this section, we propose a calibration framework with three simple yet effective strategies to alleviate this bias to achieve a more reliable and fair evaluation result.

\subsection{Multiple Evidence Calibration}
Previous studies \cite{vicuna2023,PandaLM} utilize the evaluation template that draws the conclusion first and then makes an explanation, e.g., the template used in Table \ref{tab:template}.
However, due to the nature of the auto-regressive model, the conclusions generated by the model are not supported by the explanation generated afterward.
%  post-generated
To this end, as shown in Table \ref{tab:template_cot}, 
we design an evidence calibration (EC) evaluation template $T_{EC}(Q, R1, R2)$ that requires the model to generate the explanation (evaluation evidence) first and then give the score. 
In this way, the score can be calibrated with the evaluation evidence.
To further improve the reliability of the evaluation, rather than generating only a single EC score for each response, we perform a multiple evidence calibration (MEC, Figure \ref{fig:method}(a)) that samples $k$ EC scores $\{S_{r1}^1, \dots, S_{r1}^k\}$ and $\{S_{r2}^{'1}, \dots, S_{r2}^{'k}\}$  for responses $r1$ and $r2$, where $S_r$ and $S_{r}^{'}$ denotes scores of the response $r$ at the first and second positions, respectively.
% maybe add a citation to self-consistency ? 

\subsection{Balanced Position Calibration}
We further employ a balanced position calibration (BPC) strategy to alleviate the previously identified positional bias of LLMs.
As shown in Figure \ref{fig:method}(b), for each example $(q, r1, r2)$, 
% besides the original query prompt $T_{EC}(q, r1, r2)$,
BPC additionally creates a query prompt $T_{EC}(q, r2, r1)$ by swapping the position of two responses in the original query prompt $T_{EC}(q, r1, r2)$. 
Combined with MEC, we can achieve $2k$ scores $\{S_{r1}^1, \dots, S_{r1}^k, \dots, S_{r1}^{'1}, \dots, S_{r1}^{'k}\}$ and $\{S_{r2}^{'1}, \dots, S_{r2}^{'k}, \dots, S_{r2}^1, \dots, S_{r2}^k\}$ for $r1$ and $r2$, respectively. 
The final calibrated scores of two responses ($CS_{r1}$ and $CS_{r2}$) are the average of the $2k$ scores:
\begin{equation}
    CS_{R} = \sum_{i=1}^k \frac{S_{R}^i + S_{R}^{'i}}{2k}, R=r1,r2
\end{equation}
and we regard the response with the higher average score as the better response.

\begin{table*}[t]
\centering
% \small
\scalebox{0.95}{
\begin{tabular}{llccc}
\toprule
\textbf{\textsc{Evaluators}}         & \textbf{\textsc{Methods}} & \textbf{\textsc{Accuracy}} & \textbf{\textsc{Kappa}}  & \textbf{\textsc{Cost}}\\
\midrule

{Human 1} & -  & 68.8\% & 0.50 & \$30.0 \\
{Human 2} & - & 76.3\% & 0.62 & \$30.0 \\
{Human 3} & - & 70.0\% & 0.50 & \$30.0 \\
\midrule
Human Average & - & 71.7\% & 0.54 & \$30.0 \\
\midrule
\midrule
GPT-4  & \textsc{Vanilla}  & 52.7\%  & 0.24 & \$2.00 \\
\midrule
GPT-4  & EC ($k=1$)  & 56.5\%  & 0.29 & \$2.00 \\
GPT-4  & MEC ($k=3$)   & 58.7\%  & 0.30 & \$3.19 \\
GPT-4  & MEC ($k=6$)   & 60.9\%  & 0.33 & \$6.38 \\
GPT-4  & MEC ($k=3$) + BPC ($k=3$)  & 62.5\%  & 0.37 & \$6.38 \\
GPT-4  & MEC ($k=3$) + BPC ($k=3$) + HITLC ($\beta=20\%$)  & 73.8\%  & 0.56 & \$23.1\\
\midrule
\midrule
ChatGPT  & \textsc{Vanilla}  & 44.4\% & 0.06 & \$0.10   \\
\midrule
ChatGPT  & EC ($k=1$)  & 52.6\% & 0.23 & \$0.10 \\
ChatGPT  & MEC ($k=3$)  & 53.2\% & 0.24 & \$0.17  \\
ChatGPT  & MEC ($k=6$)  & 55.6\% & 0.27 & \$0.34  \\
ChatGPT  & MEC ($k=3$) + BPC ($k=3$) & 58.7\% & 0.31 & \$0.34  \\
ChatGPT  & MEC ($k=3$) + BPC ($k=3$) + HITLC ($\beta=20\%$)  & 71.3\%  & 0.52 & \$18.3 \\
\bottomrule
\end{tabular}}
\caption{Accuracy and kappa correlation coefficient of different methods and annotators with the final voting human annotations. The \textsc{Vanilla} evaluation method was commonly used in previous works, which provided the conclusion first and then followed with the explanation. (M)EC, BPC, and HITLC denote our proposed (multiple) evidence calibration, balanced position calibration, and human-in-the-loop calibration respectively.
$\beta$\% means selecting the top-$\beta$ most likely biased examples for human annotation.
}
\label{tab:main}
\end{table*}

\subsection{Human-in-the-Loop Calibration}
\label{subsec:human-in-the-loop}
In addition to the automatic calibration strategies, another interesting question we want to explore is whether Human-In-The-Loop Calibration (HITLC) which performs the cooperation of humans and LLMs as evaluators, could stabilize the evaluation result. 
The key point of human-in-the-loop calibration is when humans should be involved in the evaluation and calibrate the evaluation result on which LLM evaluators do not perform well. 

To target the ``when'' problem, inspired by \citet{caihuman}, we introduce a \textbf{Balanced Position Diversity Entropy (BPDE)} score to find examples requiring auxiliary human calibration based on the evaluation results of MEC and BPC. 
Specifically, as shown in Figure \ref{fig:method}(c), we first compute $2k$ evaluation results $\{ \mathbf{ER}_i \}_{i=1}^{2k}$ based on the $2k$ pairs of scores.
\begin{equation}
\small
\underset{1 \le i \le k} {\mathbf{ER}_i}=\left\{
\begin{aligned}
\mathrm{\textbf{win}}, & S_{r1}^i > S_{r2}^{'i}  \\
\mathrm{\textbf{tie}}, & S_{r1}^i = S_{r2}^{'i}  \\
\mathrm{\textbf{lose}}, & S_{r1}^i < S_{r2}^{'i} 
\end{aligned}
\right.,
\underset{1 \le i \le k} {\mathbf{ER'}_i}=\left\{
\begin{aligned}
\mathrm{\textbf{win}}, &S_{r1}^{'i} > S_{r2}^{i}  \\
\mathrm{\textbf{tie}}, & S_{r1}^{'i} = S_{r2}^{i} \\
\mathrm{\textbf{lose}}, & S_{r1}^{'i} < S_{r2}^{i}
\end{aligned}
\right.,
\end{equation}
and BPDE is defined as the entropy of the evaluation results:
\begin{equation}
\mathbf{BPDE} = \sum_{\mathbf{er}\in \{ \textbf{win}, \textbf{tie}, \textbf{lose} \}} -\mathbf{p}_{\mathbf{er}} \log_{}{\mathbf{p}_{\mathbf{er}}}
\end{equation}
\begin{equation}
\mathbf{p}_{\mathbf{er}} = \frac{{\textstyle \sum_{i=1}^{k}} \mathbb{I}  (\mathbf{ER}_i = \mathbf{er}) + \mathbb{I}  (\mathbf{ER'}_i = \mathbf{er})}{2k}.
\label{equation:response_probability}
\end{equation}
A higher BPDE score indicates that  it is more likely the evaluation requires manual correction. 
A threshold is needed for BPDE as the hyper-parameter to select the top-$\beta$ most likely biased evaluations.
After selection based on the BPDE score, the annotators will evaluate the selected examples and integrate the human annotations based on the majority opinion as described in Section~\ref{subsec:human_anotation}.

\section{Experiments}
\subsection{Human Annotation}
\label{subsec:human_anotation}

To assess the effectiveness of our proposed strategies, three of the authors manually annotate the ``win/tie/lose'' outcomes of responses from ChatGPT and Vicuna-13B independently in all 80 Vicuna Benchmark questions.
All of the annotators are researchers familiar with Artificial Intelligence and are well-equipped to assess the quality of the responses.
Following the same template as the original Vicuna, the annotators are instructed to assess the responses provided by Vicuna-13B and ChatGPT from four different perspectives: helpfulness, relevance, accuracy, and level of detail. 
The responses of Vicuna and ChatGPT are presented to the annotators in random order.
The evaluation process for each example took an average of three minutes.
\textbf{The final result is based on the majority opinion among three annotators.}
\subsection{Experimental Setup and Metric}
We use the OpenAI API to conduct our experiments (``gpt-3.5-turbo-0301'' for ChatGPT, and ``gpt-4'' for GPT-4).
For the methods that do not need to sample multiple generation results, we set the generated temperature to $0$ for deterministic generation results.
For the multiple evidence strategy, we set the temperature to $1$ and sample three generation results ($k=3$). 
We use the accuracy and kappa correlation coefficient \cite{mchugh2012interrater} with the final majority of human annotation results to measure the performance of different evaluators and evaluation methods.
When calculating the results for methods that do not utilize BPC, we randomize the order of the two responses from the assistants and calculate the average results of 100 runs to ensure stable results.

\subsection{Main Results}

Table \ref{tab:main} illustrates the performance of different methods on our manually annotated $80$ annotated examples. 
As is shown:
\textbf{1)} There is a good correlation coefficient between the annotations provided by each human annotator and the final voting results. 
In detail, the average accuracy and the kappa correlation coefficient of human annotations are $71.7$\% and $0.54$, respectively;
\textbf{2)} Overall, GPT-4 achieves higher alignment with human judgments compared with ChatGPT, showing its powerful alignment ability with humans;
\textbf{3)} Compared to the commonly used \textsc{Vanilla} evaluation method, our proposed automatic calibration strategies (i.e., EC, MEC and BPC) significantly enhance the alignment between GPT-4 and ChatGPT with human judgments;
For instance, by employing the MEC and BPC calibration strategies, ChatGPT demonstrates a notable improvement in both accuracy and the kappa correlation coefficient. Specifically, the accuracy  is improved by 14.3\%, and the kappa correlation coefficient is increased from $0.06$ to $0.31$;
\textbf{4)} ``MEC ($k=3$) + BPC ($k=3$)'' outperforms ``MEC ($k=6$)'', demonstrating that LLMs are affected by positional bias, and BPC effectively ensures that LLMs serve as fair evaluators;
\textbf{5)} Our proposed HITLC can effectively enhance the alignment between GPT-4 and ChatGPT with human judgments, requiring only a small amount of human labor.
For example, by incorporating just 20\% ($\beta=20\%$) human assistance, ChatGPT attains comparable Human Average accuracy, while reducing the annotation cost from $\$30$ to $\$18.3$, a $39\%$ reduction.\footnote{The minimum hourly wage in the United States is near $\$7.5$, which can be found at \url{https://www.worker.gov/}. On average, annotating an example takes 3 minutes, and the Vicuna evaluation benchmark comprises $80$ examples in total. Consequently, the cost per annotator amounts to $\$30$. }

In conclusion, our proposed calibration methods are simple yet very effective in improving the evaluation performance with LLM as evaluators, while maintaining low costs.

\section{Analysis}
\subsection{Ablation on Evidence Number \emph{k} and Temperature \emph{t}}

\label{subsec:k_analysis}

\begin{figure}[t]
\centering
\subfigure{
\includegraphics[width=0.22\textwidth]{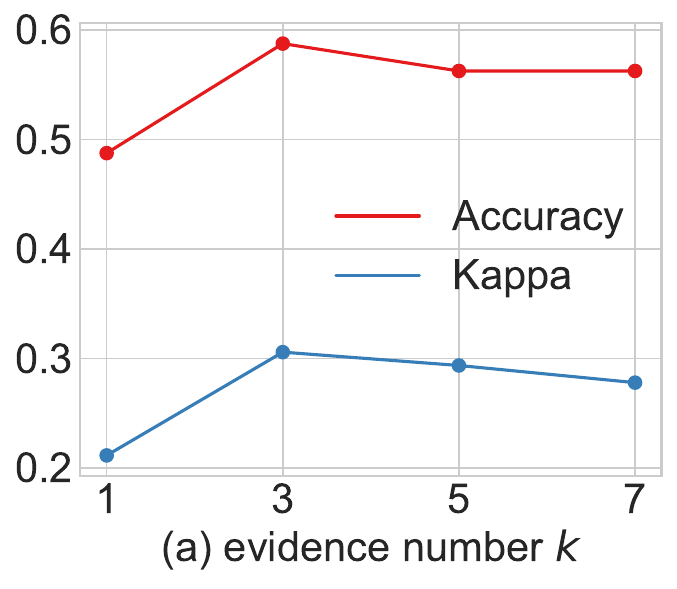}
}
\subfigure{
\includegraphics[width=0.22\textwidth]{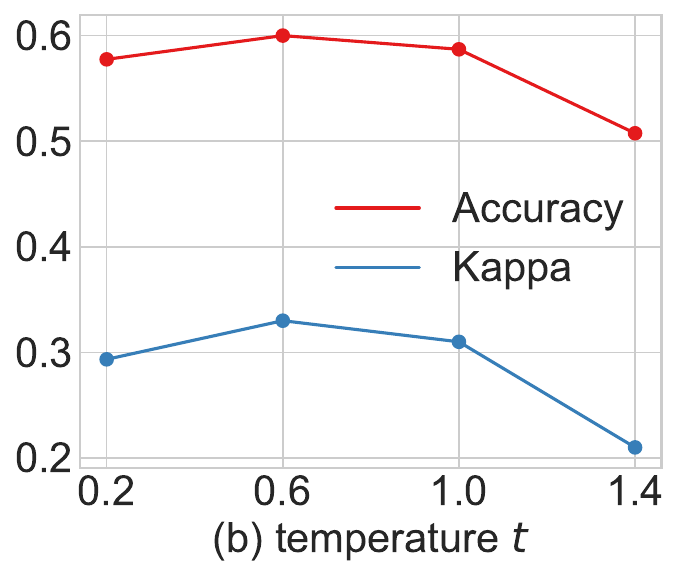}
}
\caption{Variation of accuracy and kappa coefficient with a different number of evidence $k$ and sampling temperature $t$ when ChatGPT is used as the evaluator.}
\label{fig:hyper}
\end{figure}

In the MEC and BPC strategy, we sample $k$ evaluation results for each query prompt and ensemble them to enhance the evaluation process. 
We conduct an analysis to examine the influence of the number of evidence $k$, on the model's evaluation performance. As illustrated in Figure \ref{fig:hyper}(a), we compared the performance of ChatGPT with different values of $k$, namely 1, 3, 5, and 7. The model's performance increases and then tends to be constant or decreases slightly as $k$ becomes larger. 
Despite the slight decrease, the enhancement of the model effect by the MCE strategy is still significant, illustrating the stability of the MEC strategy.
Consequently, we found that a $k$ value of $3$ yields an optimal performance.
With this value, the model achieves a notable level of performance while keeping the API cost relatively low. 

We further investigate the impact of sampling temperature $t$ on evaluation performance.
Figure \ref{fig:hyper}(b) illustrates that both low temperature (i.e., $0.2$ ) and high temperature (i.e., $1.4$) result in sub-optimal evaluation alignment. 
We believe that low temperature eliminates the randomness of sampling, weakening the effect of MEC, while high temperature compromises the quality of generation results, leading to poor performance.
Hence, it is crucial to select an appropriate temperature (e.g., $0.6$ or $1.0$ in our experiments) for the LLM evaluators.

\subsection{Effectiveness of the BPDE}

\begin{figure}[t!]
    \centering
    \includegraphics[width=0.9\linewidth]{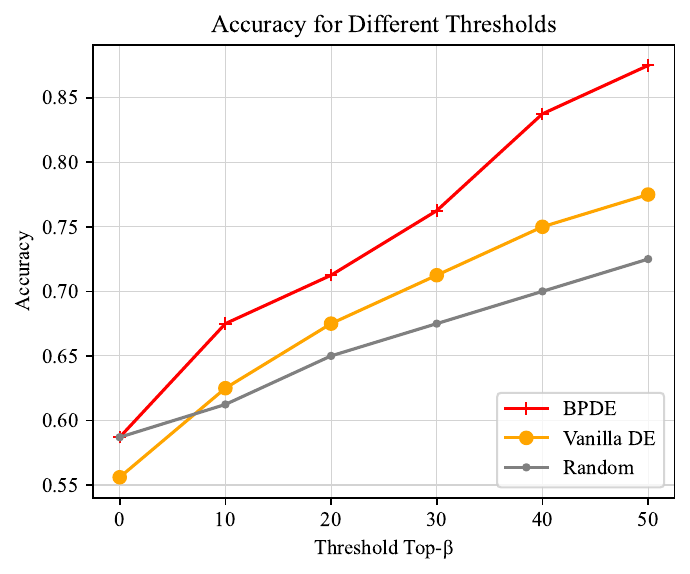}
    \caption{The accuracy of various methods changes with different human assistant thresholds (Top-$\beta$) when ChatGPT is used as the evaluator.}
    \label{fig:HITLC_line}
\end{figure}

Our HITLC strategy utilizes BPDE score to select examples for human annotations.
In order to analyze the efficiency of BPDE score, 
we compare BPDE with two typical baselines, \textit{Random} and \textit{Vanilla Diversity Entropy}, where Random denotes randomly select examples for human annotations, and Vanilla Diversity Entropy is calculated by using only the evaluation results of one position without swapping the position of two responses. 
To ensure fairness, the total number of evaluation results is $6$ for both BPDE and Vanilla Diversity Entropy.
As shown in Figure \ref{fig:HITLC_line}:
\textbf{1)} Two Diversity Entropy methods outperform Random, showing the effectiveness of selecting examples based on the diversity entropy;
\textbf{2)} BPDE outperforms Vanilla DE, which shows LLMs are sensitive to position exchange, and the results of BPC can significantly improve the performance of HITLC compared to relying solely on the results of MEC.

% \subsection{Case Study}

\subsection{Generalization on the Pairwise Comparison Evaluation Template}

\begin{table}[t]
\centering
\scalebox{0.83}{
\begin{tabular}{llccc}
\toprule
\textbf{\textsc{Templates}}         & \textbf{\textsc{Methods}} & \textbf{\textsc{Acc.}} & \textbf{\textsc{Kap.}}  & \textbf{\textsc{C.R}}\\
\midrule

\textsc{Scoring}  & \textsc{Vanilla}  & 44.4\% & 0.06 & 82.5\%   \\
\midrule
\textsc{Scoring}  & MEC  & 53.2\% & 0.24 & 35.0\%   \\
\textsc{Scoring}  & MEC + BPC & 58.7\% & 0.31 & N/A   \\

\midrule
\midrule
\textsc{Comparing}  & \textsc{Vanilla}  & 50.2\% & 0.18 & 50.0\%  \\
\midrule
\textsc{Comparing}  & MEC & 54.8\% &  0.27 & 42.5\%  \\
\textsc{Comparing}  & MEC + BPC & 60.3\% & 0.35 &  N/A \\

% \midrule

\bottomrule
\end{tabular}}
\caption{Effectiveness of our proposed two automatic calibrated methods on two different evaluation templates with ChatGPT as the evaluator. \textsc{Acc.}, \textsc{Kap.} and \textsc{C.R} are short for Accuracy, Kappa correlation coefficient, and Conflict Rate, respectively. N/A means the Conflict Rate is not valid for BPC methods.}
\label{tab:other-template}
\end{table}

\begin{figure*}[!t]
\centering
\subfigure{
\includegraphics[width=0.48\textwidth]{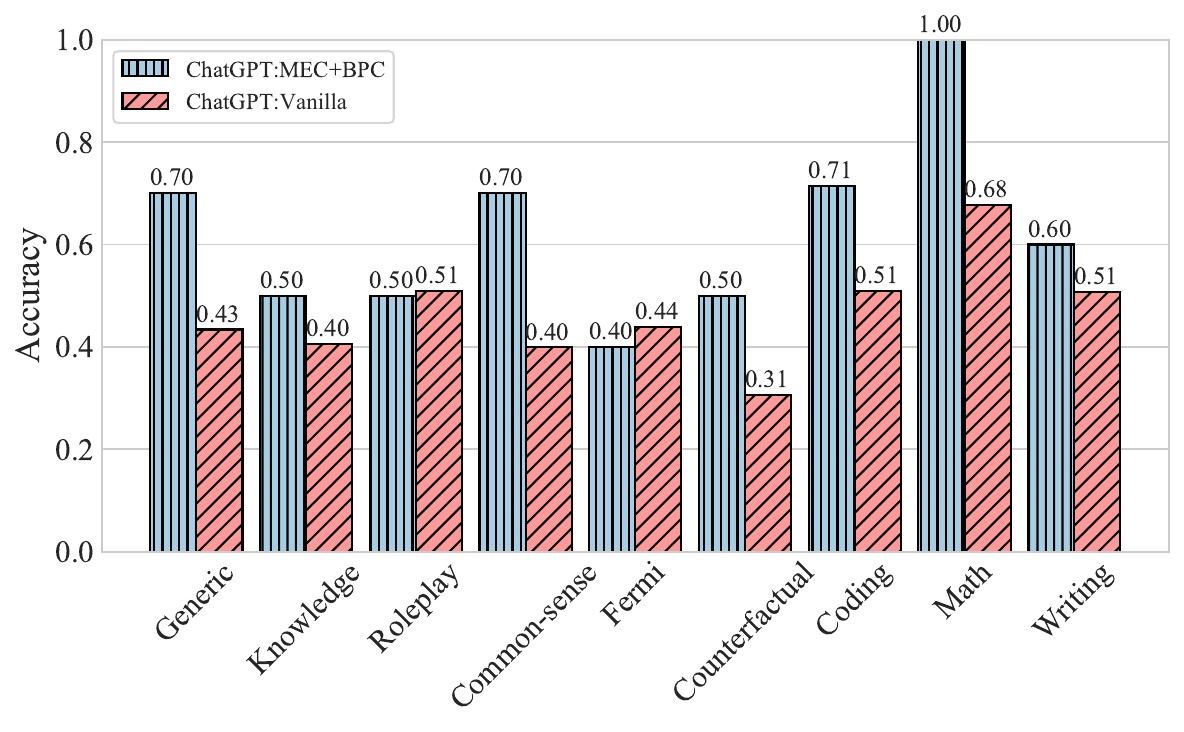}
}
\subfigure{
\includegraphics[width=0.48\textwidth]{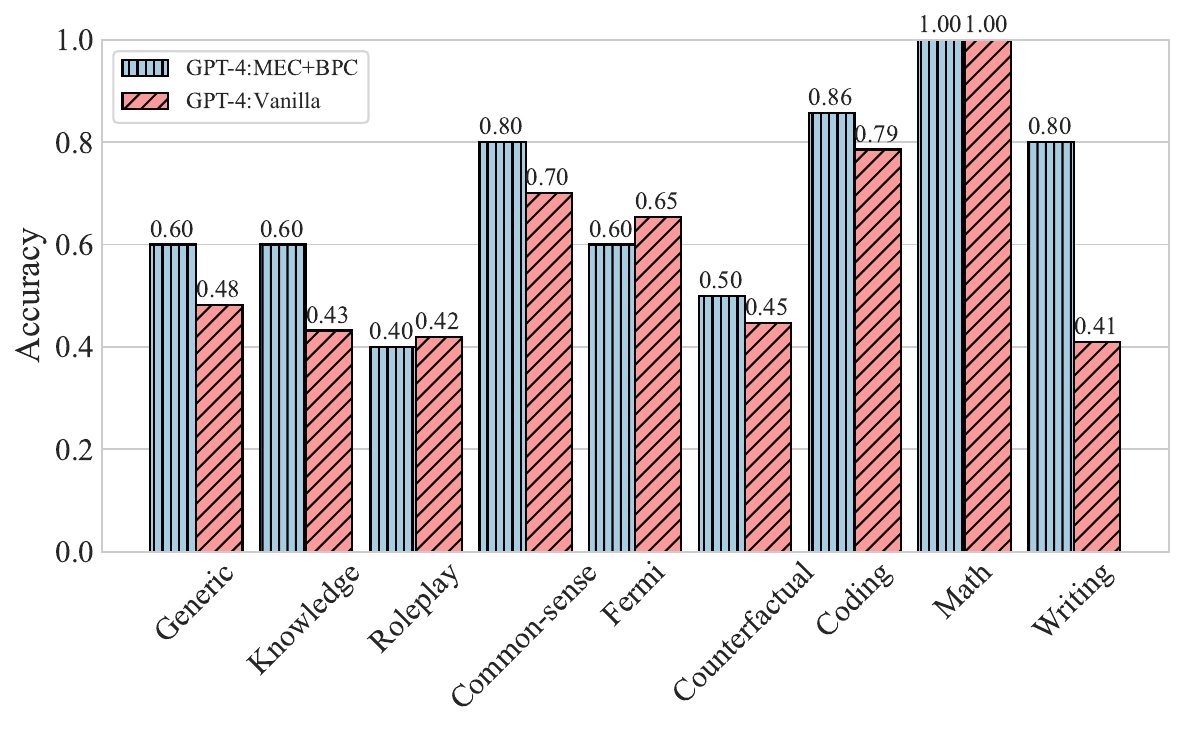}
}
\caption{Fine-grained analysis of evaluation quality. Our MEC and BPC improve the evaluation performance of ChatGPT and GPT-4 in nearly all categories. Especially on the complex task categories such as common sense, coding, and math for ChatGPT. }
\label{fig:fine-grained}
\end{figure*}

To provide a more comprehensive validation of our proposed calibration methods, in addition to the previous \textsc{Scoring} evaluation template that rates each response, 
we extend our analysis to incorporate the \textsc{Comparing} evaluation template. This template facilitates a direct comparison between two responses, eschewing explicit scores in its assessment.
Specifically, we prompt LLMs to produce results labeled as ``Assistant 1'', ``Assistant 2'', or ``Same'', indicating whether the response from Assistant 1 is better, worse, or equal to that of Assistant 2.
As is shown in Table \ref{tab:other-template}:
\textbf{1)} Our proposed methods are applicable to both of these templates, leading to enhanced accuracy and a heightened correlation coefficient for ChatGPT;
\textbf{2)} The significant performance gap (nearly 6\% accuracy) between the \textsc{Vanilla} method of two templates, coupled with the high conflict rate, highlights the sensitivity and unreliability of LLMs. 
However, our methods effectively narrow this performance gap and reduce conflict, showcasing how calibration enhances LLM robustness.

\subsection{Fine-Grained Analysis of Evaluation Quality}
\label{subsec:fine_aspect}

In order to further analyze the evaluation capabilities of the model, we perform a fine-grained analysis of the questions by dividing them into $9$ categories following \citet{vicuna2023}.
We calculate the performance of different evaluators within these categories. 
As shown in Figure \ref{fig:fine-grained}, we find that: 
\textbf{1)} In certain complex tasks such as common-sense, coding and math, GPT-4 performs significantly better than ChatGPT, highlighting the strength of GPT-4 as a more fair evaluator in these scenarios;
\textbf{2)} Our proposed MEC+BPC strategy demonstrates noticeable improvement in evaluating ChatGPT's performance on complex tasks, allowing us to obtain satisfactory evaluation results with a low API cost.

\section{Related Work}
\subsection{Evaluation of Large Language Models}
LLMs have demonstrated powerful general generation capabilities, becoming universal assistants \cite{chatgpt, gpt4, song2023restgpt}.
With the rapid advancement of LLMs, it becomes crucial to evaluate their ability to follow human instructions.
Traditional evaluation methods assess the ability by calculating a metric, like BLEU, ROUGE, BERTScore, or BARTScore, to compare the generated response with a reference response.
However, these metrics do not adequately measure the alignment of the generated response with human intent \cite{he-etal-2023-blind}.
While human evaluation is treated as the most accurate measurement of model performance, it is costly and time-consuming to operate at scales.
Considering the potent capabilities of LLMs, researchers have started utilizing LLMs to evaluate the proficiency of generative models in adhering to human instructions \cite{vicuna2023,lu2023error,li2023m}.
In these works, Vicuna's evaluation paradigm \cite{vicuna2023} is widely adopted, where it provides a question and two responses from two models, and uses GPT-4 to determine which response has better quality.

\subsection{Bias of Deep Neural Networks}
Deep Neural Networks have been proven to easily learn biases from the data, which significantly impacts their reliability.
Specifically, bias has also been investigated in natural language inference ~\citep{gururangan2018annotation, mccoy-etal-2019-right, belinkov-etal-2019-dont,liu-etal-2020-hyponli,liu-etal-2020-empirical}, question answering~\citep{min-etal-2019-compositional}, ROC story cloze~\citep{DBLP:conf/acl/CaiTG17, DBLP:conf/conll/SchwartzSKZCS17}, lexical inference \cite{levy2015supervised}, visual question answering \cite{goyal2017making}, information extraction \cite{wang2021behind,aca,song2023repcl,xia-etal-2023-enhancing} and so on.
LLMs are pre-trained using a vast amount of data from the internet, making it highly likely for them to learn biases present in those materials.
Although the LLMs are already widely adopted as a proxy of human evaluators, the reliability of this paradigm is not well explored.
In this paper, we critically examine the LLMs-as-evaluator paradigm and uncover a significant positional bias.
Furthermore, we propose three simple yet effective methods to calibrate the positional bias to achieve reliable and fair evaluation results.

\section{Conclusion}
In this paper, we reveal a systematic positional bias in evaluation with advanced ChatGPT/GPT-4 models: by manipulating the order of candidate responses during evaluation, the quality ranking results can be significantly influenced. 
To this end, we introduce three effective strategies, namely Multiple Evidence Calibration (MEC), Balanced Position Calibration (BPC), and Human-in-the-Loop Calibration (HITLC).
MEC requires the LLM evaluator to first provide multiple evaluation evidence to support their subsequent ratings and BPC aggregates the results from various orders to determine the final score. 
Based on the results of MEC and BPC, HITLC further calculates a balanced position diversity entropy to select examples for human annotations.
These strategies successfully reduce the evaluation bias and improve alignment with human judgments. 
We provide our code and human annotations to support future studies and enhance the evaluation of generative models.

\bibliography{FairEval}
\bibliographystyle{acl_natbib}

\clearpage

\end{document}